\documentclass[10pt,twocolumn,letterpaper]{article}

\usepackage{iccv}
\usepackage{times}
\usepackage{epsfig}
\usepackage{graphicx}
\usepackage{amsmath}
\usepackage{amssymb}
\usepackage{eucal}
\usepackage{mathtools} 
\usepackage{yhmath}
\usepackage[acronym]{glossaries}
\usepackage{color, colortbl}
\usepackage{enumitem}

\makenoidxglossaries
\newacronym{ml}{ML}{Machine Learning}
\newacronym{dl}{DL}{Deep Learning}
\newacronym{dgm}{DGM}{Deep Generative Model}
\newacronym{rl}{RL}{Reinforcement Learning}
\newacronym{dnn}{DNN}{Deep Neural Network}
\newacronym{cnn}{CNN}{Convolutional Neural Network}
\newacronym{vae}{VAE}{Variational Autoencoder}
\newacronym{gan}{GAN}{Generative Adversarial Network}
\newacronym{nf}{NF}{Normalizing Flow}
\newacronym{dm}{DM}{Diffusion Model}
\newacronym{ae}{AE}{Autoencoder}
\newacronym{mse}{MSE}{Mean Squared Error}
\newacronym{lvm}{LVM}{Latent Variable Models}
\newacronym{ebm}{EBM}{Energy-Based Model}
\newacronym{ddpm}{DDPM}{Denoising Diffusion Probabilistic Model}
\newacronym{dsm}{DSM}{Denoising Score Matching}
\newacronym{sde}{SDE}{Stochastic Differential Equation}
\newacronym{ncsn}{NCSN}{Noise Conditional Score Network}
\newacronym{sgd}{SGD}{Stochastic Gradient Descent}
\newacronym{mcmc}{MCMC}{Markov Chain Monte Carlo}
\newacronym{ddim}{DDIM}{Denoising Diffusion Implicit Model}
\newacronym{mlp}{MLP}{Multilayer Perceptron}
\newacronym{3dmm}{3DMM}{3D Morphable Face Model}
\newacronym{vqgan}{VQGAN}{Vector-Quantized Generative Adversarial Network}
\newacronym{nlp}{NLP}{Natural Language Processing}
\newacronym{eer}{EER}{Equal-Error Rate}
\newacronym{cpd}{CPD}{Contextual Partial Dropout}
\newacronym{auc}{AUC}{Area Under Curve}
\newacronym{fdr}{FDR}{Fisher Discriminancy Ratio}
\newacronym{fr}{FR}{Face Recognition}
\newacronym{frm}{FRM}{Face Recognition Model}
\newacronym{adagn}{AdaGN}{Adaptive Group Normalization}
\newacronym{silu}{SiLU}{Sigmoid Linear Unit}
\newacronym{ffhq}{FFHQ}{Flickr-Faces-HQ}
\newacronym{lfw}{LFW}{Labeled Faces in the Wild}
\newacronym{calfw}{CA-LFW}{Cross-Age Labeled Faces in the Wild}
\newacronym{cplfw}{CP-LFW}{Cross-Pose Labeled Faces in the Wild}
\newacronym{cfpfp}{CFP-FP}{Celebrities in Frontal-Profile in the Wild}
\newacronym{agedb30}{AgeDB-30}{AgeDB with 30-year age gap protocol}
\newacronym{ldm}{LDM}{Latent Diffusion Model}
\newacronym{ca}{CA}{Cross-Attention}
\newacronym{ema}{EMA}{Exponential Moving Average}

\newacronym{fmr}{FMR}{False Match Rate}
\newacronym{fnmr}{FNMR}{False Non-Match Rate}
\newacronym{far}{FAR}{False Acceptance Rate}
\newacronym{frr}{FRR}{False Rejection Rate}
\newacronym{fmr100}{FMR100}{\acrshort{fnmr} at a \acrshort{fmr} of $1\%$}
\newacronym{fmr1000}{FMR1000}{\acrshort{fnmr} at a \acrshort{fmr} of $0.1\%$}
\newacronym{roc}{ROC}{Receiver Operating Characteristic}

\newacronym{sota}{SOTA}{state-of-the-art}
\newacronym{gdpr}{GDPR}{General Data Protection Regulation}

\newacronym{rq}{RQ}{Research Question}

\newcommand{\cacpdzeroshort}{CPD0 }
\newcommand{\cacpdtwentyfiveshort}{CPD25 }
\newcommand{\cacpdfiftyshort}{CPD50 }

\newcommand{\cacpdzeroshortwithoutspace}{CPD0}
\newcommand{\cacpdtwentyfiveshortwithoutspace}{CPD25}
\newcommand{\cacpdfiftyshortwithoutspace}{CPD50}

\newcommand{\approachname}{IDiff-Face }
\newcommand{\approachnamewithoutspace}{IDiff-Face}

\newcommand{\ra}[1]{\renewcommand{\arraystretch}{#1}}
\usepackage{rotating}

\DeclarePairedDelimiterX{\infdivx}[2]{(}{)}{%
  #1\;\delimsize\|\;#2%
}

\DeclarePairedDelimiter{\norm}{\lVert}{\rVert} 
\usepackage{pifont}
\newcommand*{\FeatureTrue}{\ding{52}}
\newcommand*{\FeatureFalse}{\ding{56}}

\definecolor{Gray}{gray}{0.95}

\usepackage[pagebackref=true,breaklinks=true,letterpaper=true,colorlinks,bookmarks=false]{hyperref}
\usepackage{tabularx}   
\usepackage{booktabs}   

\iccvfinalcopy 

\def\iccvPaperID{10991} 

\ificcvfinal\fi

\usepackage[accsupp]{axessibility}  

\usepackage{physics}
\usepackage{amsmath}
\usepackage{bm}
\usepackage{tikz}
\usepackage{mathdots}
\usepackage{yhmath}
\usepackage{cancel}
\usepackage{color}
\usepackage{xcolor}
\usepackage{siunitx}
\usepackage{array}
\usepackage{multirow}
\usepackage{amssymb}
\usepackage{gensymb}
\usepackage{tabularx}
\usepackage{extarrows}
\usepackage{booktabs}
\usepackage{scalerel}
\usepackage{hyperref}
\usepackage{caption}
\usepackage{subcaption}
\usepackage{amsthm}

\usepackage[symbol]{footmisc}

\begin{document}

\title{IDiff-Face: Synthetic-based Face Recognition through Fizzy Identity-Conditioned Diffusion Models}

\author{Fadi Boutros$^{1}$, Jonas Henry Grebe$^{1}$,  Arjan Kuijper$^{1,2}$, Naser Damer$^{1,2}$\\
$^{1}$Fraunhofer Institute for Computer Graphics Research IGD, Darmstadt, Germany\\
$^{2}$Department of Computer Science, TU Darmstadt,
Darmstadt, Germany\\
Email: fadi.boutros@igd.fraunhofer.de
}

\maketitle
\ificcvfinal\thispagestyle{empty}\fi

\begin{abstract}
The availability of large-scale authentic face databases has been crucial to the significant advances made in face recognition research over the past decade. However, legal and ethical concerns led to the recent retraction of many of these databases by their creators, raising questions about the continuity of future face recognition research without one of its key resources. Synthetic datasets have emerged as a promising alternative to privacy-sensitive authentic data for face recognition development. However, recent synthetic datasets that are used to train face recognition models suffer either from limitations in intra-class diversity or cross-class (identity) discrimination, leading to less optimal accuracies, far away from the accuracies achieved by models trained on authentic data. This paper targets this issue by proposing IDiff-Face, a novel approach based on conditional latent diffusion models for synthetic identity generation with realistic identity variations for face recognition training. Through extensive evaluations, our proposed synthetic-based face recognition approach pushed the limits of state-of-the-art performances, achieving, for example, $98.00\%$ accuracy on the \acrfull{lfw} benchmark, far ahead from the recent synthetic-based face recognition solutions with $95.40\%$ and bridging the gap to authentic-based face recognition with $99.82\%$ accuracy\footnote[1]{\url{https://github.com/fdbtrs/idiff-face}}. 
\end{abstract}


\vspace{-5mm}
\section{Introduction}
\vspace{-2mm}
\acrfull{fr} is one of the most widely used biometric technologies due to the high accuracies achieved by the recent \acrshort{fr}s \cite{AdaFace2022,CosFace,ElasticFace} with a wide range of applications such as logical access control to portable devices \cite{prakash2021biometric,DBLP:journals/access/BoutrosSKDKK22}. 
This ubiquitous adoption has been fuelled by the application of deep learning to \acrshort{fr} and the rapid research advances in this direction, mainly on novel margin-penalty based softmax losses \cite{ArcFace, ElasticFace} and deep network architectures \cite{ResNet,DBLP:journals/access/BoutrosSKDKK22}. However, this rapid progress has only been possible due to the public availability of large-scale \acrshort{fr} training databases \cite{VGGFace2, MSCeleb1MDataset}.
Such databases contain millions of images, and they are typically collected from the internet without proper user consent, which raises concerns about the legal and ethical use of these databases for \acrshort{fr} development.

Since, for example, the European Union (EU) adopted the \acrfull{gdpr} \cite{european_commission_regulation_2016} in 2018, more justified criticism of the associated privacy risks has been raised against the usage of public biometric datasets that were collected without proper consent. The \acrshort{gdpr} explicitly grants individuals the "right to be forgotten" and enforces stricter requirements on the collection, distribution, and usage of face databases, making it extremely hard, or even infeasible, to maintain such regulations. Thus, many of them \cite{VGGFace2, UMDFacesDataset, MSCeleb1MDataset} that are widely used to train \acrshort{fr}s were retracted by their creators to avoid legal complications, which raises the question about continuity of \acrshort{fr} research since the availability of one of its key resources became questionable.

As an effort to address these legal and ethical concerns, synthetic data has recently emerged as a promising alternative to authentic databases for \acrshort{fr} training \cite{Qiu2021, Boutros2022SFace, FBoutros2022USynthFace, IVC_Syn, DigiFace1M, QuantFace}. 
This is also the trend for FR subsystems, such as morphing and spoof attack detection \cite{DBLP:conf/cvpr/DamerLFSPB22,DBLP:conf/cvpr/FANG23} and face images quality assessment \cite{DBLP:journals/corr/abs-2305-05768}.
This research direction is driven by the progress on \acrlong{dgm}s (\acrshort{dgm}s), a model designed to learn the probability distribution of a certain dataset, enabling the generation of completely new synthetic samples. 
This process can be conditioned on specific attributes such as age, facial expression, and a defined set of visual appearances, e.g. the lighting condition and head pose \cite{Deng2020, Shoshan2021, Tewari2020, Ghosh2020}. 
Most of the \acrshort{dgm} approaches for generating synthetic faces are based on \acrlong{gan}s (\acrshort{gan}s) \cite{Goodfellow2014, Deng2020, Shen2020, IDnet, Boutros2022SFace}. A number of recent works \cite{Marriott2020, FBoutros2022USynthFace, Qiu2021} utilized GANs \cite{Deng2020,Karras2020StyleGANADA} to generate synthetic data for \acrshort{fr} training. The reported results by \acrshort{sota} synthetic-based \acrshort{fr} showed significant degradation in the verification accuracies in comparison to \acrshort{fr} trained on authentic data. This performance gap is mainly due to low identity discrimination \cite{Boutros2022SFace} or small intra-class variation \cite{FBoutros2022USynthFace,Qiu2021} in their training datasets. A realistic trade-off between these two properties, as we will demonstrate in this paper, is needed to achieve high verification accuracies.

Recently, \acrlong{dm}s (\acrshort{dm}s) \cite{Ho2020, Nichol2021, Rombach2021, Dhariwal2021} gained attention for both research and industry due to their potential to rival \acrshort{gan}s on image synthesis \cite{Dhariwal2021}, even though they are easier to train, have no stability issues, and stem from a solid theoretical foundation. Besides that, they can be conditioned on additional information, and demonstrated impressive results in a wide range of tasks, especially in text-to-image generation such as 
OpenAI's DALL-E 2 \cite{Ramesh2022}, Stable Diffusion \cite{Rombach2021}, and Google's Imagen \cite{Saharia2022}. 

\begin{figure}[t!]
    \begin{center}
        \includegraphics[width=0.9\linewidth]{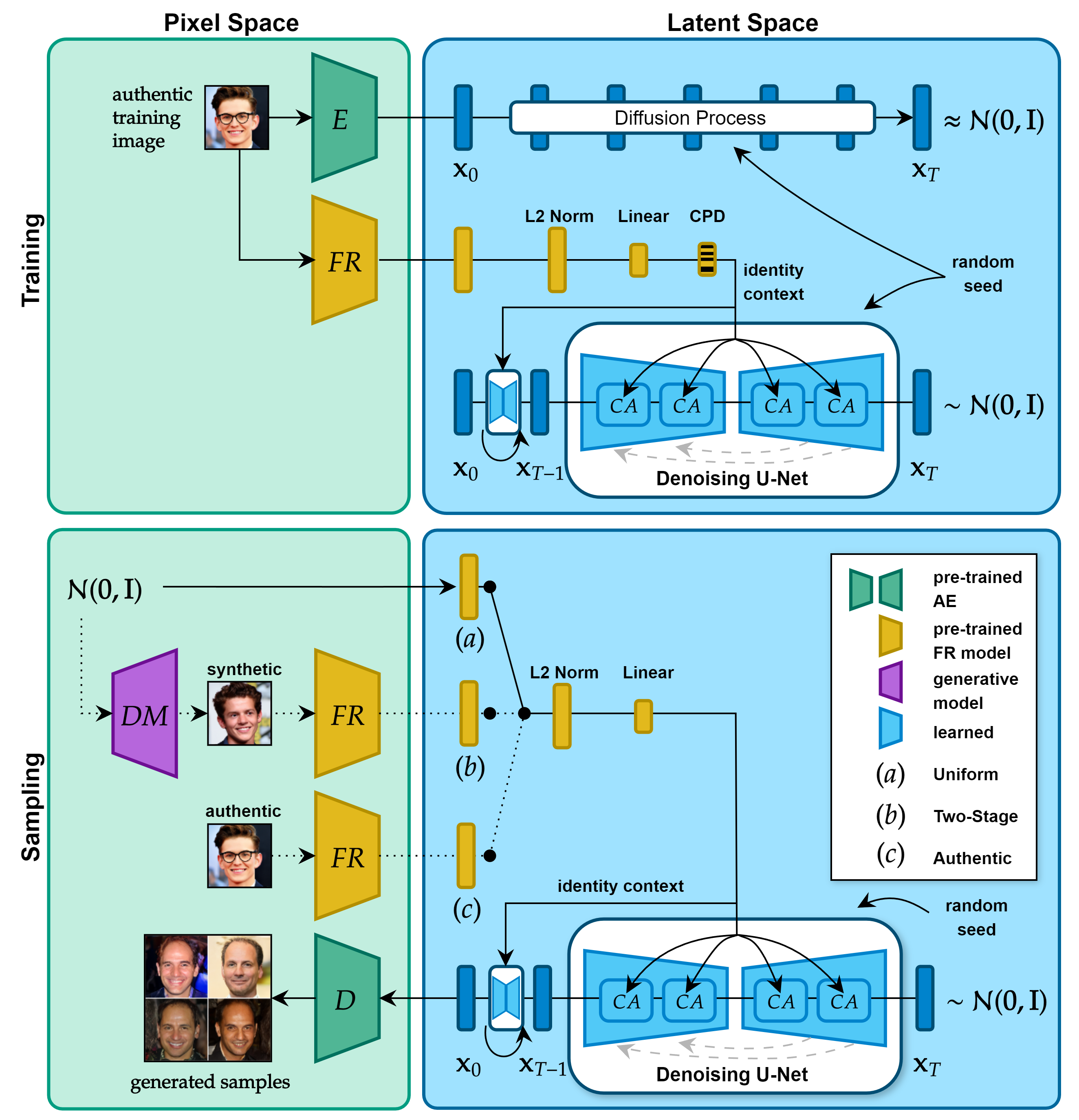}
    \end{center}
    \vspace{-6mm}
    \caption{Overview of the proposed \approachnamewithoutspace. This diagram is divided into two parts, the upper part  visualizes the training procedure, and the lower part  shows the conditional sampling process, partially inspired by \cite{Rombach2021}. \textbf{Top}: During training, the learned denoising U-Net is conditioned on a context that is based on the feature representation obtained through a pre-trained FR model. The entire DM training process is in the latent space of a pre-trained AE.
    The diffusion process that basically provides the targets for learning the iterative reverse process is depicted above. \textbf{Bottom}: For samples generation, the trained DM can generate samples based on three types of identity contexts:  authentic, two-stage, or synthetic uniform representations. By fixing the identity context and varying the added noise, different samples for the same identity can be generated. 
    }
    \label{fig:approach_overview}
    \vspace{-5mm}
\end{figure}

We propose in this work an identity-conditioned \acrlong{dm} approach, namely \approachnamewithoutspace. Our \approachname is designed and trained to generate synthetic images of synthetic identities that are identity-separable, with a desirable relatively large intra-class diversity.
Our approach follows the common concept \cite{Rombach2021} of dividing the training into two stages by first training an \acrshort{ae} (or using a pre-trained \acrshort{ae}) and then leveraging the resulting representational low-dimensional latent space of the \acrshort{ae} for \acrshort{dm} training \cite{Rombach2021}.  The identity condition is introduced to our \approachname by projecting the training images into a low-dimensional feature representation and then injecting it into the \acrshort{dm}'s intermediate representations through a \acrfull{ca} mechanism \cite{Rombach2021}. To ensure that our synthetic data contains, to a large degree, realistic variations and to avoid overfitting our \approachname to the information encoded in the identity context, we proposed a simple, yet effective \acrfull{cpd} approach that partially drops out components with a certain probability of the identity context during the training phase, thus the term "fizzy" in our title. 
We first demonstrate the identity discrimination and intra-class variation of our synthetically generated data. We also compared our dataset in terms of identity-separability and intra-class variation, with the recent \acrshort{sota} synthetic datasets. 
As we empirically present in this paper, the synthetic datasets that are used in \acrshort{sota} synthetic-based \acrshort{fr} training maintain, to some degree, identity discrimination, however, only with a low intra-class variation or vice versa. 
Unlike these approaches, our proposed \approachname offers a more realistic trade-off between identity discrimination and intra-class variation that is controlled by \acrshort{cpd}.
Achieving this realistic trade-off between these two properties is necessary to achieve verification accuracy that is close to verification accuracies achieved by \acrshort{fr}s trained on authentic data as authentic data naturally contains such properties. By utilizing our synthetic data ($500$K images) for \acrshort{fr} training and under the same training setups and dataset size, our synthetic-based \acrshort{fr} achieved an average accuracy of $88.20\%$, significantly outperforming 
all \acrshort{sota} synthetic-based \acrshort{fr}s (best average accuracy was $83.45\%$) and closing the gap to \acrshort{sota} \acrshort{fr} trained on authentic data, where the average accuracy on CASIA-WebFace \cite{Yi2014} ($500$K images) and MS1MV2 \cite{MSCeleb1MDataset,ArcFace} ($5.8$M images) were $94.92\%$ and $97.18\%$, respectively.
Our model also outperformed human-level performance in face verification, where the reported accuracy on \acrfull{lfw} \cite{LFWDatabase} was $97.5\%$ \cite{DBLP:conf/iccv/KumarBBN09} and our model achieved $98.00\%$.

\vspace{-3mm}
\section{Related Work}
\vspace{-1mm}
The vast majority of \acrshort{dgm}s that are proposed in the literature are often designed for generating synthetic images of random identities \cite{Goodfellow2014,Karras2017,Karras2018,IVC_Syn} or editing certain facial attributes of existing reference images \cite{Tran2019,Shen2018FaceFeatGAN,Shen2020,Tewari2020}, and thus, such models are not by design capable of generating synthetic identities with multiple different images per identity.
While other \acrshort{dgm}s proposed to generate images of synthetic identities, most of them either rely on pre-existing semantic attribute annotations \cite{Tran2019,Shen2018,Shen2018FaceFeatGAN,Shen2020,Deng2020,Kowalski2020,Ghosh2020,Shoshan2021,Tewari2020}, meticulously constructed training batches \cite{Donahue2017,Deng2020,Shoshan2021}, or the supervision by sophisticated \acrlong{3dmm}s (\acrshort{3dmm}) \cite{Shen2018, Shen2018FaceFeatGAN,Deng2020,Kowalski2020,Ghosh2020,Tewari2020}. In some cases, even all of these requirements are necessary in order to explicitly model specific parametric factors, such as age or illumination, and thus, gain control over the generative process \cite{Deng2020}. 
These approaches presented impressive results in generating high-quality and realistic images \cite{Deng2020,Shoshan2021} with an unprecedented level of control. However, the visual appearances in their generated images are limited to a predefined set of attributes \cite{Deng2020,Shoshan2021}, and thus, their synthetic images might not contain natural real-world variations or large diversities in terms of utility \cite{DBLP:conf/iwbf/FuKBD23}. Utilizing such synthetic data for \acrshort{fr} training might lead to suboptimal verification accuracies. For that reason, SynFace \cite{Qiu2021} proposed a synthetic-based \acrshort{fr} approach based on DiscoFaceGAN \cite{Deng2020} with synthetic identity mix-up to enhance the intra-class diversity. 
USynthFace \cite{FBoutros2022USynthFace} proposed the use of unlabelled synthetic data for unsupervised \acrshort{fr} training, achieving competitive accuracies.
During \acrshort{fr} training, USynthFace \cite{FBoutros2022USynthFace} proposed to utilize intensive data augmentation, which significantly improved the overall verification accuracy.
On the other hand, SFace \cite{Boutros2022SFace} and IDnet \cite{IDnet} proposed to train a StyleGAN-ADA \cite{Karras2020StyleGANADA} under a class-conditional setting. 
SFace \cite{Boutros2022SFace} images contain a higher intra-class variation, which, comes at the cost of low identity discrimination, when compared to the other approaches \cite{FBoutros2022USynthFace}. 
ExFaceGAN \cite{ExFaceGAN} presented a framework to disentangle identity information in
learned latent spaces of GANs to generate multiple samples of any synthetic identity. 
DigiFace-1M \cite{DigiFace1M} utilized a digital rendering pipeline to generate synthetic images based on a learned model of facial geometry and a collection of textures, hairstyles, and 3D accessories. DigiFace-1M \cite{DigiFace1M} also applied aggressive data augmentations for their \acrshort{fr} training. However, it comes at a considerable computational cost during the rendering process. 
All previously mentioned approaches presented promising accuracies. Nonetheless, these accuracies are still significantly lower than the ones achieved by \acrshort{fr} trained on authentic data, where for example, the average accuracy (on five benchmarks, see Table \ref{tab:frm_training_validation_large_scale}) of \acrshort{sota} synthetic-based \acrshort{fr} was $83.45\%$, far away from $94.82\%$ achieved by authentic-based \acrshort{fr}.

Motivated by the remarkable text-to-image results of recent approaches that leverage \acrshort{dm}s \cite{Rombach2021, Ramesh2022, Saharia2022} and in an effort to bridge the gap between synthetic- and authentic-based \acrshort{fr} performance, this paper is the first to propose an identity-conditioned approach based on \acrshort{dm} to generate synthetic identity-specific images with a more realistic intra-class diversity for \acrshort{fr} training, outperforming all recent \acrshort{sota} synthetic-based \acrshort{fr}s with an obvious margin.

\vspace{-3mm}
\section{Methodology }
\vspace{-2mm}
This section describes our proposed \approachnamewithoutspace approach to generate identity-specific yet realistic synthetic face images. \approachname is based on a \acrshort{dm} that is conditioned on identity contexts. During the training stage, our proposed \approachname is conditionally trained on authentic embeddings obtained from the authentic training dataset. After training, our \approachname can be used either to generate variations of existing authentic images by using authentic embeddings or to generate novel synthetic identities by using synthetic embeddings. In Figure \ref{fig:approach_overview}, an overview of the proposed method is provided, with a clear distinction between the training and the sampling stages. In order to generate samples of synthetic identities, a synthetic identity representation has to be created, e.g. synthetic uniform or synthetic two-stage contexts, as explored in this work.
\vspace{-1mm}
\subsection{Identity-Conditioned Latent Diffusion}
\vspace{-1mm}
Our \approachname is based on a \acrshort{ddpm} that is trained in the latent space of a pre-trained \acrshort{ae} \cite{Rombach2021} and conditioned on identity-contexts i.e. feature representations extracted using a \acrshort{fr} model.
A \acrshort{ddpm} \cite{Ho2020} is a \acrshort{dm} that discretizes the diffusion processes into a finite number $T$ of steps and learns to reverse this process by training a conditional \acrshort{dnn} to estimate the noise that has been added to a sample $\mathbf{x}_t$ at time step $t$. The architecture of our \approachname is a modified U-Net \cite{Ronneberger2015} based on \cite{Ho2020} that includes residual and attention blocks.
We incorporate the identity context of a sample $\mathbf{x}$ in the \acrshort{dm} to encourage the \acrshort{dm} to learn to generate identity-specific face images. 
This has been achieved by mapping $\mathbf{x} \in \mathbb{R}^{W\times H\times C}$ into a feature representation $f(\mathbf{x}) = \mathbf{c} \in \mathbb{R}^{d}$ using a pre-trained face recognition model $f$ optimized to learn discriminant identity information, which is then injected to the \acrshort{dm} using \acrshort{ca} mechanism as proposed by \cite{Rombach2021}. An ablation study on two different conditional mechanisms, including conditional \acrshort{ca} \cite{Rombach2021} and \acrfull{adagn} \cite{Dhariwal2021} for identity-specific images generation is provided in the supplementary material.

Let $0<\beta_1,\beta_2,...,\beta_T<1$ be a fixed (linear) variance schedule and $\mathbf{x}_0$ be the encoded image $\mathbf{x}$ in the pre-trained latent space. The learnable \textit{reverse diffusion process} is defined as a Markov chain with Gaussian transitions $p_\theta(\mathbf{x}_{t-1}|\mathbf{x}_t)$ starting at a prior distribution $p(\mathbf{x}_T)=\mathcal{N}(\mathbf{0,I})$. As proposed by the \acrshort{ddpm} authors, this transition kernel is parameterized by the \acrshort{dnn} with parameters $\theta$ that learn to predict the noise at time step $t$ based on the current estimate $\mathbf{x}_t$ and, in our case, the additional condition $\mathbf{c}$. With the notational abbreviations $\alpha_t = 1-\beta_t$ and $\overline{\alpha}_t := \sum_{i=1}^t \alpha_i$ \cite{Ho2020}, our conditional variant of the original \acrshort{ddpm} training objective is given as:
\vspace{-1mm}
\begin{align*}
\small
    \mathcal{L}(\theta) &:= \mathbb{E}_{t,\mathbf{x}_t, \boldmath{\epsilon}} \left [\norm{\boldmath{\epsilon} - \boldmath{\epsilon}_\theta(\mathbf{x}_t,t,\mathbf{c})}^2_2\right ]\\
    &= \mathbb{E}_{t,\mathbf{x}_0, \boldmath{\epsilon}} \left [ \norm{\boldmath{\epsilon} - \boldmath{\epsilon}_\theta ( \sqrt{\overline{\alpha}_t}\mathbf{x}_0 + \sqrt{1-\overline{\alpha}_t} \boldmath{\epsilon}, t,\mathbf{c})  }^2_2 \right].
\end{align*}

This expectation can be minimized through \acrshort{sgd} by randomly drawing samples $(\mathbf{x}_0,t,\boldmath{\epsilon},\mathbf{c})$ with
$\mathbf{x}_0\in \mathcal{D}, t\sim\mathcal{U}(1,T)$, $\boldmath{\epsilon} \sim \mathcal{N}(\mathbf{0, I})$, and $\mathbf{c}=f(\mathbf{x})$  and then minimizing the \acrshort{mse} between the true and the predicted noise.

Analogously, the original \acrshort{ddpm} \cite{Ho2020} sampling is slightly modified to incorporate identity contexts $\mathbf{c}$ into the iterative sampling process, which mirrors a score-based sampling chain with \textit{Langevin dynamics} using additionally added noise vectors $\boldmath{\zeta}_t\sim \mathcal{N}(\mathbf{0,I})$ at each time step $t$.
\vspace{-1mm}
\begin{align*}
\small
    \mathbf{x}_{t-1} &= \boldmath{\mu}_\theta(\mathbf{x}_t,t,\mathbf{c}) + \sigma_t \boldmath{\zeta}_t \\
&= \frac{1}{\sqrt{\alpha_t}} \left ( \mathbf{x}_{t} - \frac{1-\alpha_t}{\sqrt{1-\overline{\alpha}_t}} \epsilon_\theta(\mathbf{x}_t, t, \mathbf{c}) \right ) + \sigma_t \boldmath{\zeta}_t .
\end{align*}
At the end of the iterative process, the final estimate of $\mathbf{x}_0$ from the last iteration has to be mapped back from the latent space to the pixel space by using the pre-trained decoder.
\vspace{-1mm}
\subsection{Synthetic Image Generation}
\vspace{-1mm}
As the proposed \approachname is conditionally trained on identity contexts, i.e feature representations of input face images, such feature representations are also required to generate synthetic samples. Feature representations $\mathbf{c}$ can be obtained from any authentic image $\mathbf{x}$ and, in this case, the \approachname will generate variations of $\mathbf{x}$ using randomly sampled $\mathbf{x}_T \sim \mathcal{N}(\mathbf{0,I})$ and different seeds for the non-deterministic sampling process. To generate images of synthetic identities, we explored two different approaches. First, one can randomly sample the identity context from a uniform distribution over the hypersphere, because face representations are typically L2-normalized \cite{ElasticFace,CosFace}. In order to generate samples from this distribution, a naive way of doing it would be to simply generate a random direction from a uniform distribution $\mathcal{U}(-1,1)$ for each of the components before normalizing the resulting vector to get a point on the surface of the sphere. Unfortunately, this does not result in a uniform distribution over the surface. Instead, the components of the initial direction have to be sampled from a spherical Gaussian distribution, e.g. $\mathcal{N}(\mathbf{0},\mathbf{I})$, to obtain a uniform distribution after the normalization process \cite{Marsaglia1972}. After sampling, one can utilize it as a fixed input to our \approachname to generate different samples from the same identity by varying the noise seeds. 
The second approach is to utilize an additional unconditional \acrshort{dm} to generate a synthetic image $\mathbf{x}'$. After that, a feature representation $\mathbf{c}'$ can be extracted using $f$ as $\mathbf{c}'=f(\mathbf{x}')$. Analogously to the first approach, different samples of the same identity can be obtained by fixing $\mathbf{c}'$ throughout different samples.

The latter approach will be referred to as Two-Stage \approachnamewithoutspace, where a random synthetic image from a random identity is generated first using an unconditional \acrshort{dm} model and then it is used as input to our conditional \approachname to generate different samples of the same identity.

\begin{figure}[!]
\begin{center}
   \includegraphics[width=0.95\linewidth]{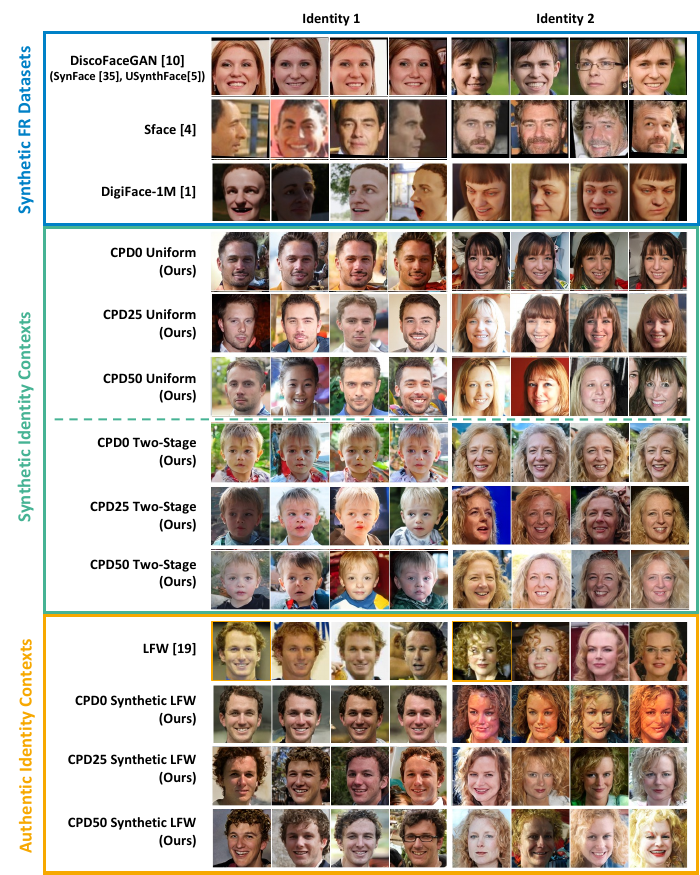}
\end{center}
\vspace{-5mm}
   \caption{Samples from recent synthetic \acrshort{fr} training datasets and our \approachname. The upper box (in blue) shows samples from the three \acrshort{dgm}s that are used in the \acrshort{sota} synthetic-based \acrshort{fr}s. The next group (in green) presents samples from our \approachname models with different \acrshort{cpd} probabilities and different types of synthetic embeddings. The last group (in yellow) demonstrates the generation of variations for existing \acrshort{lfw} identities, whose reference images are framed in yellow colour. There are four consecutive images per identity, and two identities in total are presented for each method. Zoom in for the best view.}
\label{fig:sample_overview}
\vspace{-5mm}
\end{figure}

\vspace{-1mm}
\subsection{Enhancing Intra-Class Variation via \acrshort{cpd}}
\vspace{-2mm}
To prevent the model from overfitting to identity contexts, which limits the intra-class variation in the generated samples, we introduced Dropout \cite{Srivastava2014} to the \approachname training. If the network is overfitted to the identity-context condition, the generated images using the same identity context will be almost identical i.e. small intra-class variations, regardless of the initial starting noise or random seed. 
This has been achieved by dropping out components of a context embedding, each with a certain probability. This method is referred to as \acrfull{cpd} as the context is only partially dropped out during training.

\vspace{-3mm}
\section{Experimental Setup }
\vspace{-1mm}
\subsection{\approachname Training Dataset}
\vspace{-1mm}
The proposed \acrshort{ddpm} was trained using the \acrshort{ffhq} dataset \cite{Karras2018}, which consists of $70{,}000$ high-quality images of human faces, showcasing a diverse range of attributes such as age, lightning, and facial expressions.
The face images used for training have a resolution of $128\times 128$ pixels. 

\vspace{-2mm}
\subsection{\approachname Training Setups}
\vspace{-1mm}
The experiments in this paper were conducted on a cluster of $8$ NVIDIA A100-SXM4-40GB GPUs. The proposed $\epsilon$-prediction model (U-Net) \cite{Rombach2021} takes a three-channel input and $96$ channels for the initial image projection layer, which are then used as input for the first residual block \cite{ResNet}. The network has four resolution levels, and the multipliers for the number of channels used on those levels are $1$, $2$, $2$, and $2$ respectively. Attention mechanisms use a fixed number of $32$ channels per head, with the number of heads calculated based on the number of incoming channels. Attention blocks are applied in all residual blocks except the first resolution level \cite{Rombach2021}. Each resolution level has $2$ consecutive residual blocks. For the experiments, three different levels of \acrshort{cpd} probability have been explored: $0\%$, $25\%$, or $50\%$. We trained and evaluated three instances of our \approachname with these three dropout probabilities, which will be denoted as \cacpdzeroshortwithoutspace, \cacpdtwentyfiveshortwithoutspace, and \cacpdfiftyshortwithoutspace.

The training process follows a batch-wise training loop over the training dataset $\mathcal{D}$ with half-precision (float16) and a learning rate schedule based on the number of global steps. An annealing cosine learning rate schedule with warm restarts \cite{CosineAnnealing} is used, which reduces the learning rate in repeating phases where the first phase is $10{,}000$ steps long and each subsequent phase is twice as long as the previous one. Therefore, the number of global steps per training is set to $150{,}000$ steps, which corresponds to $4$ full phases. Also, the Adam optimizer \cite{AdamOptimizer} is used with an initial learning rate of $\gamma=1e-4$. Following \cite{Ho2020, Rombach2021}, we apply an \acrshort{ema} to the weights of the model with a negative exponential factor of $0.75$, leading to an effective decay factor of $0.999$ at $10{,}000$ steps. Therefore, a second \acrshort{ema} copy of the currently trained model is maintained for inference purposes. 
The batch size is fixed at $512$ and is equally split across $8$ GPUs. 
The training dataset is augmented with horizontal flipping with a $50\%$ probability.
Regarding the diffusion process, $T=1{,}000$ time steps and a linear diffusion variance schedule are used throughout all experiments.


\textbf{Autoencoder:} The latent space of the pre-trained \acrshort{ae} is used to lower the computational demands and facilitate the learning process by learning a perceptually similar but less complex space for the \acrshort{dgm} before the \acrshort{ldm} is trained. For the \acrshort{ae}, a pre-trained \acrshort{vqgan} model (\textit{vq-f4} first stage model) from the official \acrshort{ldm} \cite{Rombach2021} repository is used. A \acrshort{vqgan} can be understood as a regularized \acrshort{ae}, which enforces a discrete latent space by learning a codebook of latent vector and replacing the encoder's outputs with its nearest-neighbor codes during inference. During training, \acrshort{vqgan}s are guided by an additional patch-based adversarial objective $\mathcal{L}_{\text{adv}}$ to better match the realism of the training data with the reconstructions. The pre-trained \acrshort{vqgan} provides an effective trade-off between sample compression and generative abilities as demonstrated in recent work \cite{Rombach2021}.

\textbf{Identity Context:} The identity context is obtained by projecting the training image into its feature representation using a \acrshort{sota} pre-trained \acrshort{fr}, namely ElasticFace \cite{ElasticFace}. We used the official pre-trained model released by \cite{ElasticFace}. 
The network architecture of ElasticFace \cite{ElasticFace} is ResNet-100 \cite{ResNet} trained on the MS1MV2 \cite{MSCeleb1MDataset,ArcFace} dataset using ElasticFace-Arc loss and the output feature dimensionality is $512$.

\begin{figure}[h]
\begin{center}

  \includegraphics[width=0.30\linewidth]{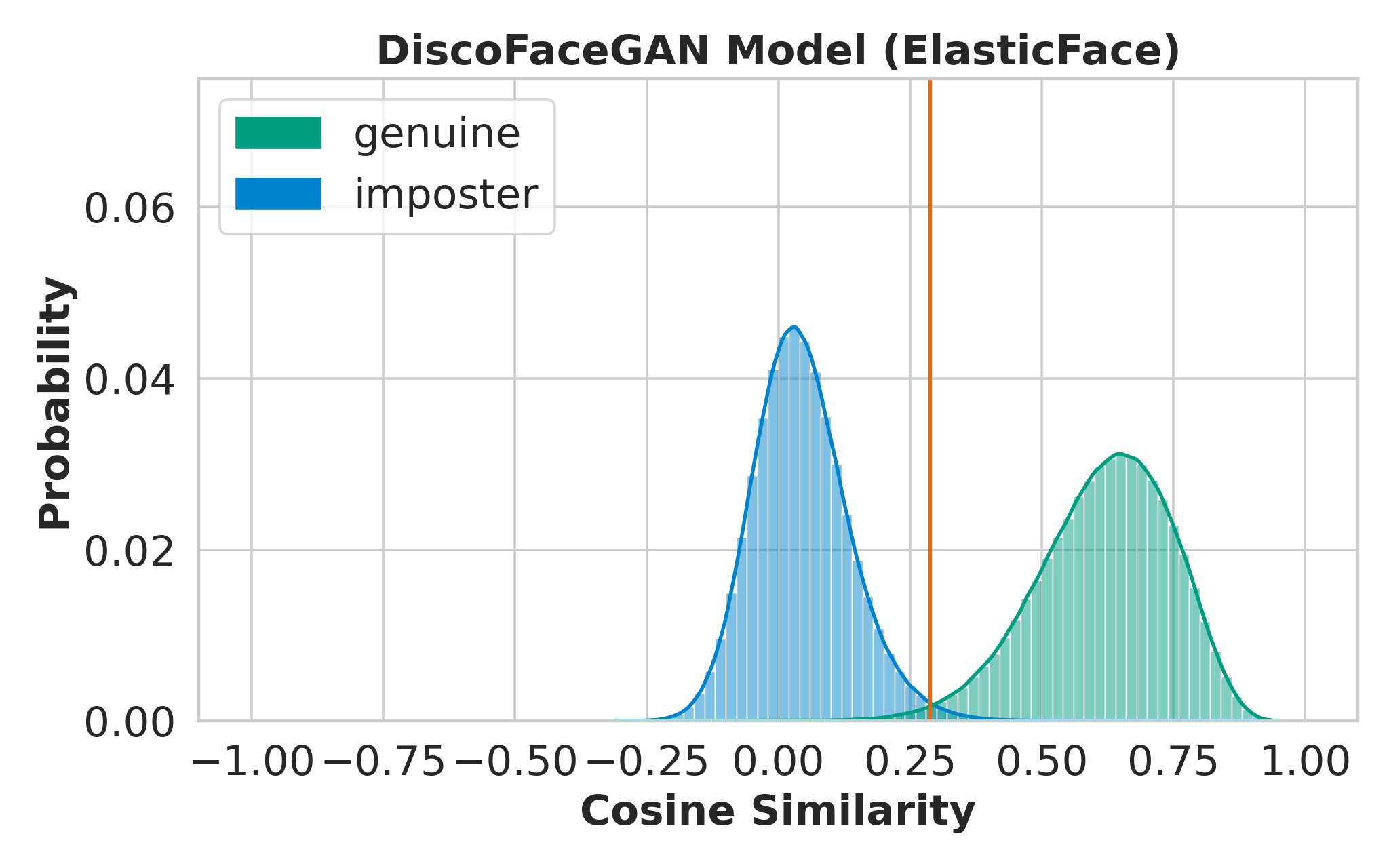}
  \includegraphics[width=0.30\linewidth]{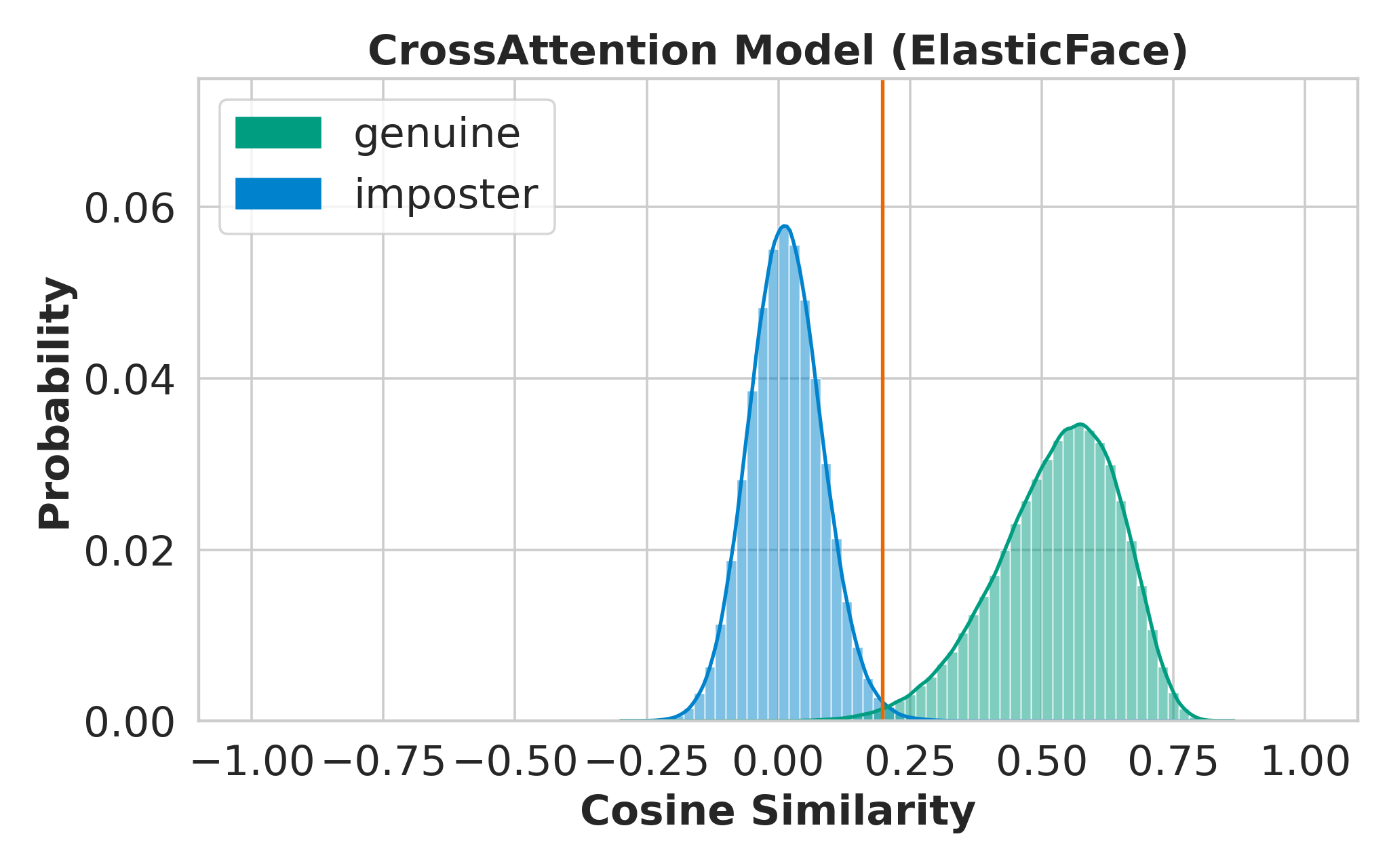}
  \includegraphics[width=0.30\linewidth]{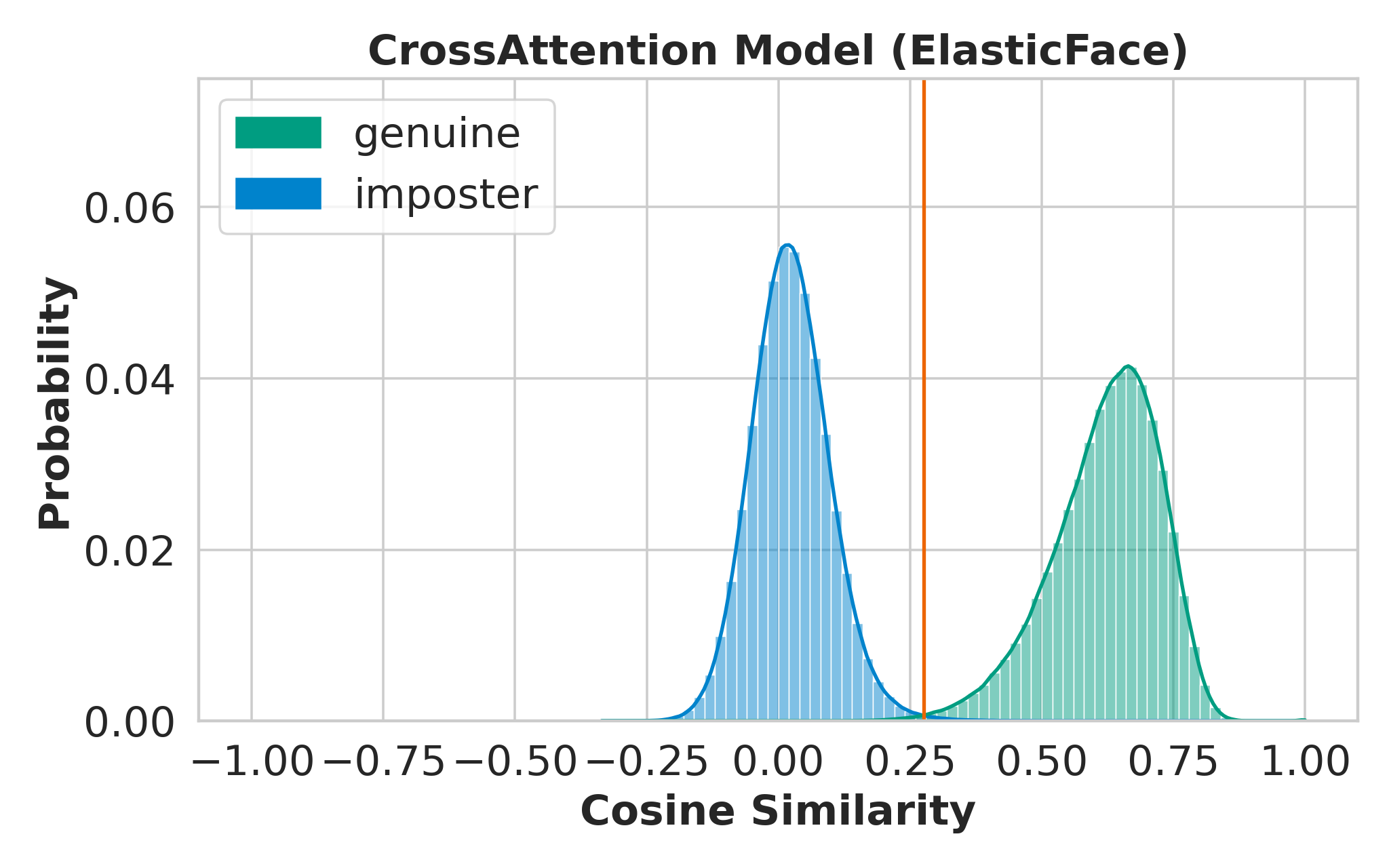}

  \includegraphics[width=0.30\linewidth]{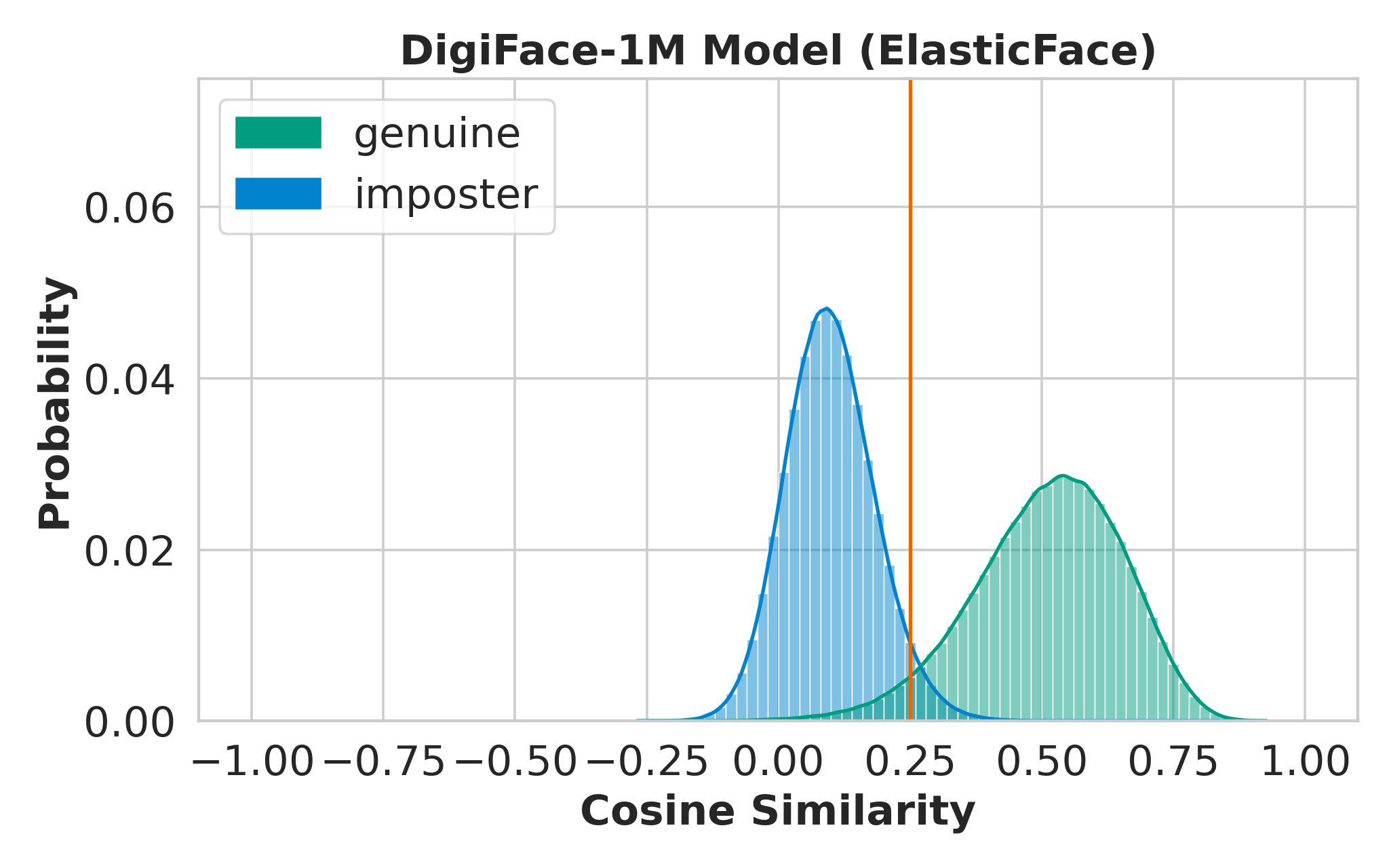}
  \includegraphics[width=0.30\linewidth]{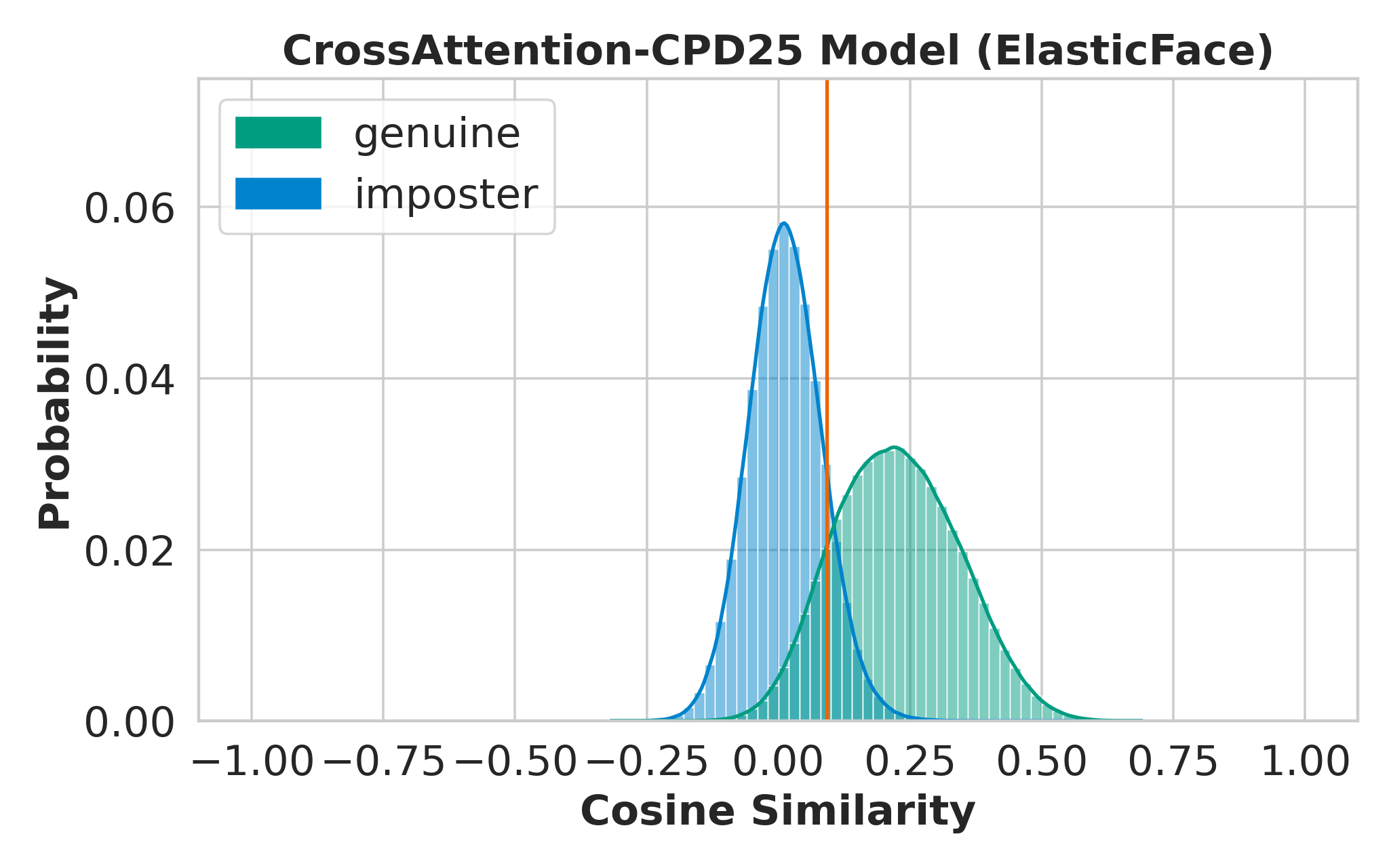}
  \includegraphics[width=0.30\linewidth]{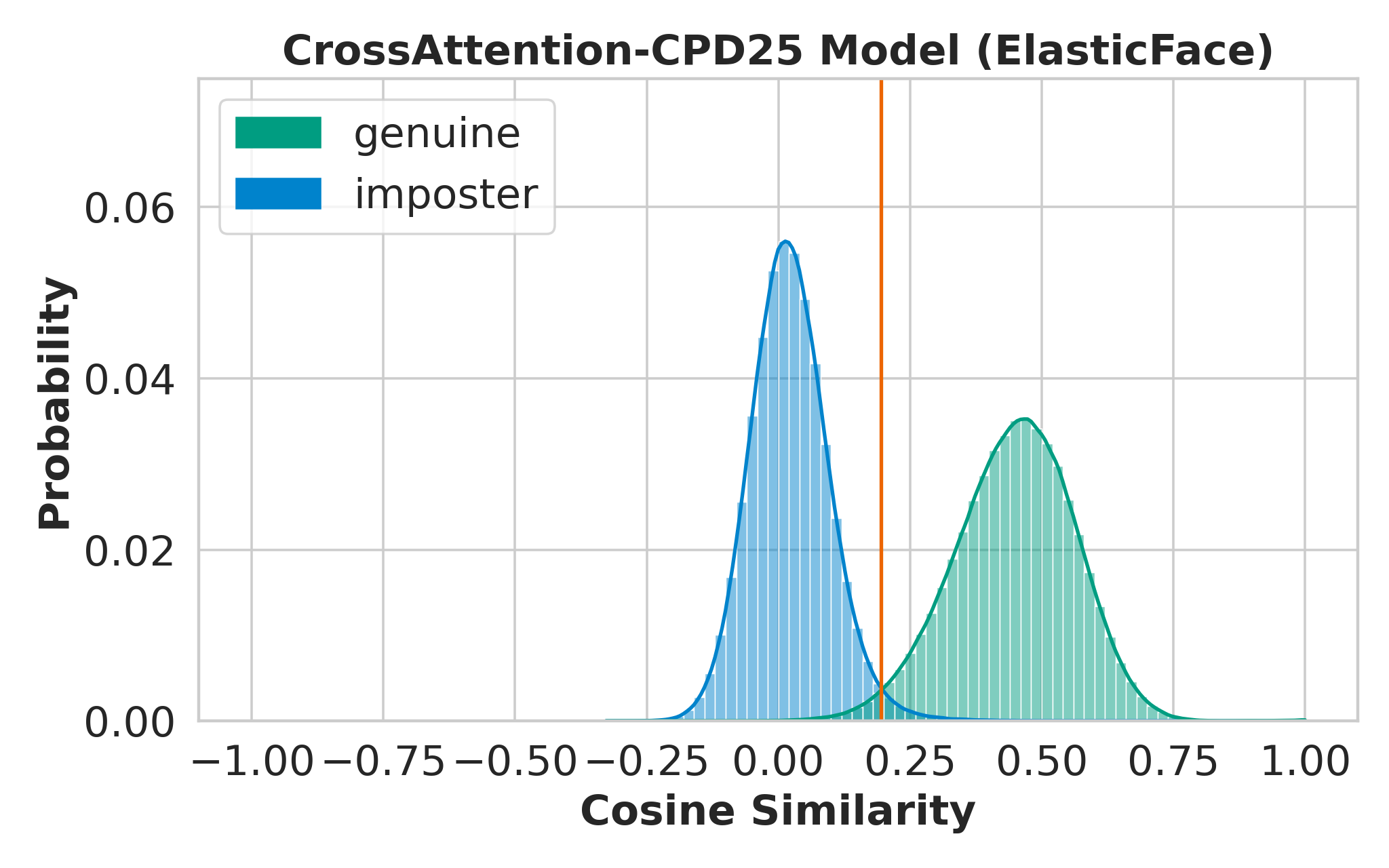}

  \includegraphics[width=0.30\linewidth]{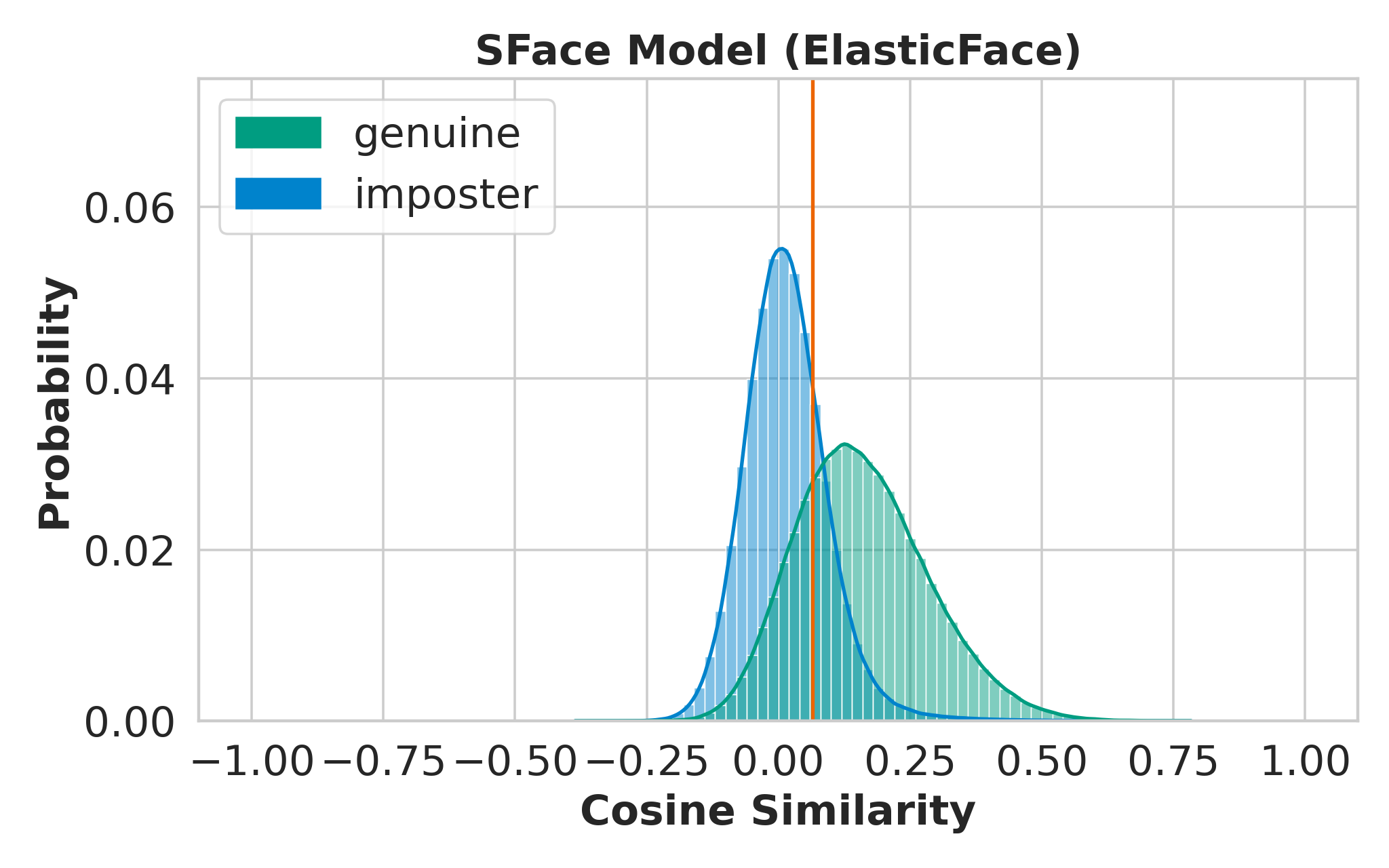}
  \includegraphics[width=0.30\linewidth]{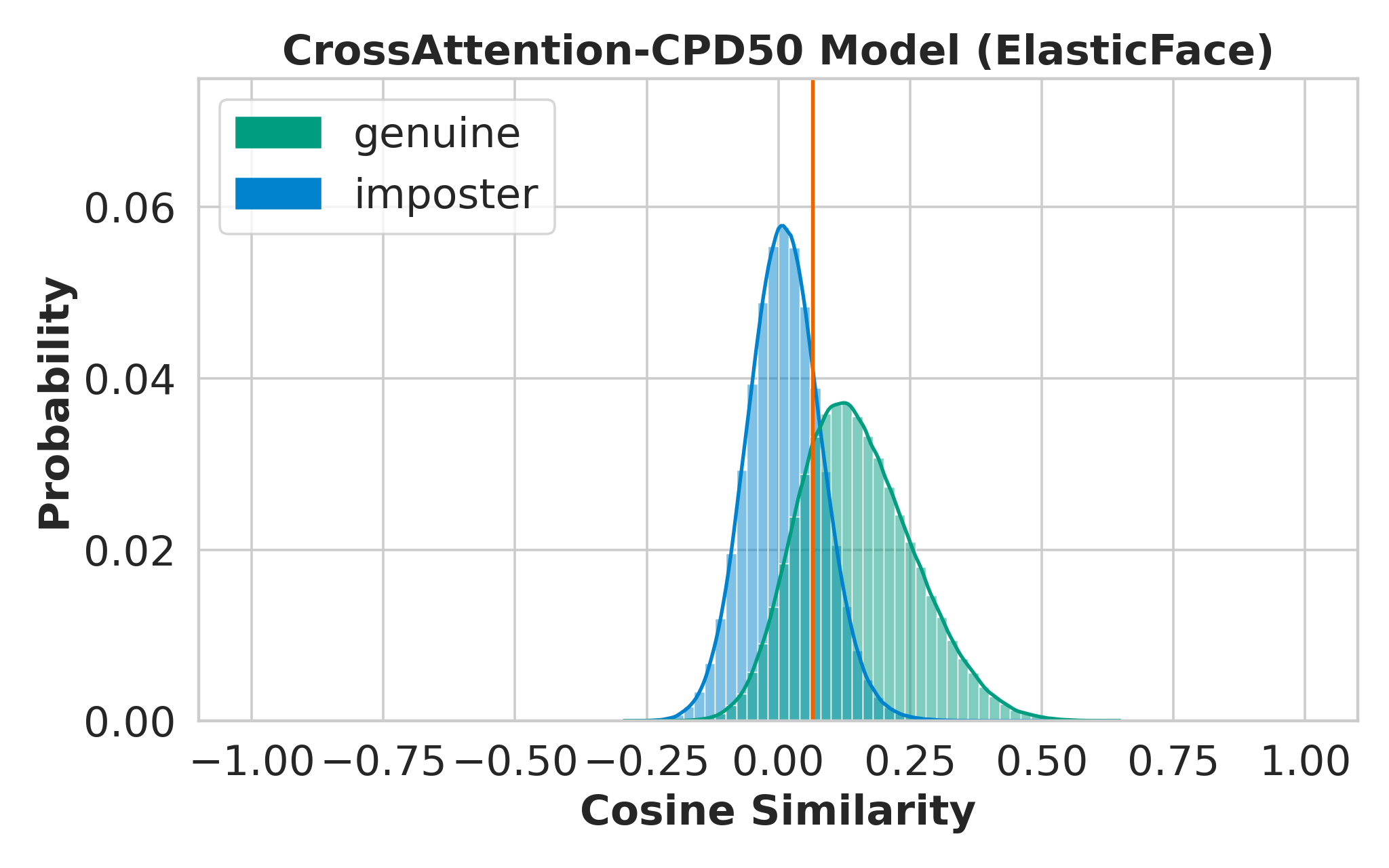}
  \includegraphics[width=0.30\linewidth]{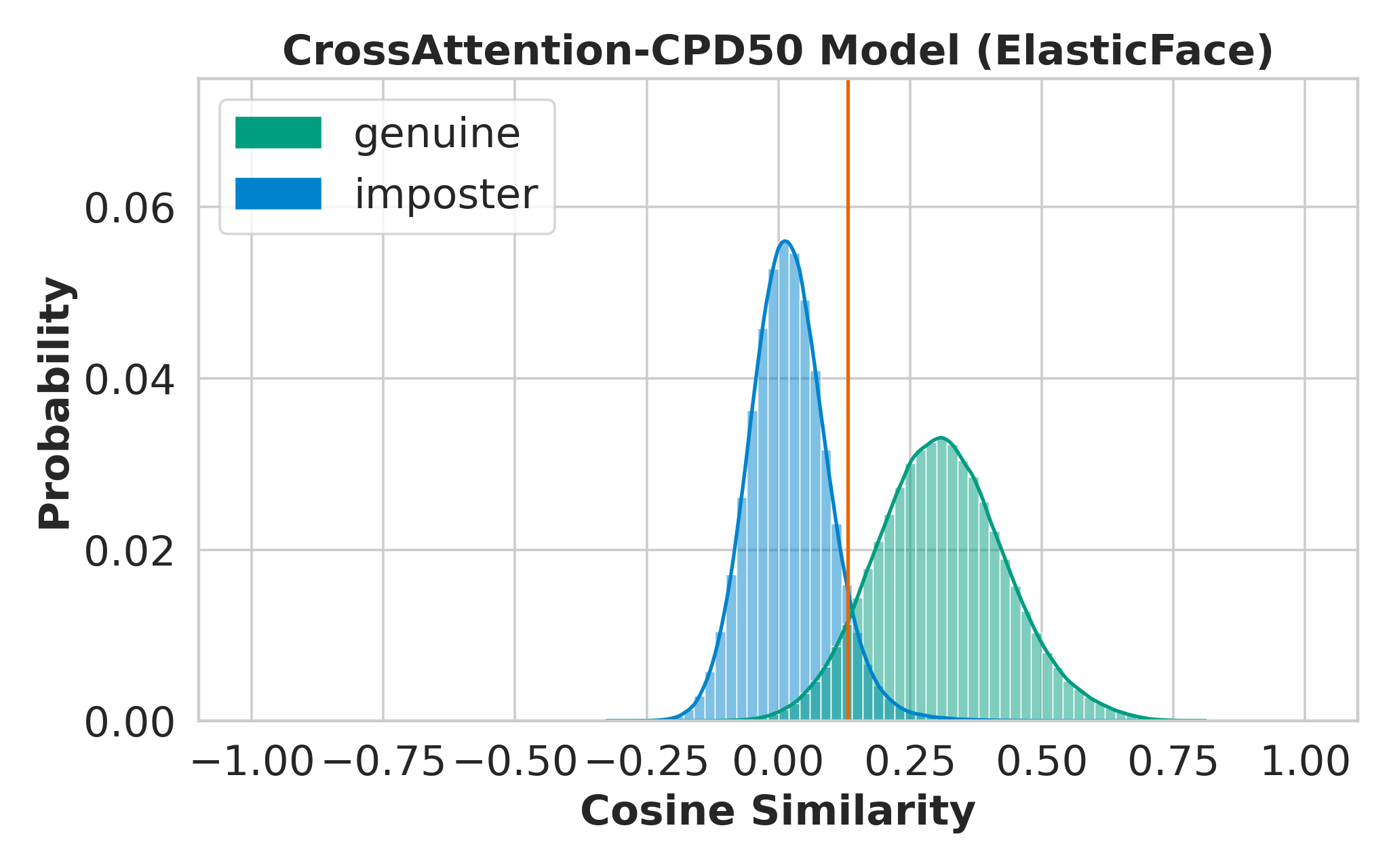}
\end{center}
   \vspace{-5mm}
   \caption{Syn-vs-syn genuine and imposter comparison score distributions for different \acrshort{dgm}s. The first column presents the distributions obtained from samples of \acrshort{dgm}s used in \acrshort{sota} synthetic-based \acrshort{fr}s. The second column presents the results from our three models based on synthetic uniform embeddings, while the last column shows our distributions based on synthetic two-stage embeddings. For each, all genuine comparison scores have been computed ($\approx1{,}200{,}000$ pairs) based on $5$ K identities with $16$ images each, and the same number of imposter scores has been randomly sampled.}
\label{fig:syn_vs_syn_distributios_overview}
\vspace{-3mm}
\end{figure}

\vspace{-4mm}
\section{Results}
\vspace{-2mm}

\textbf{Assessment of Intra-Class Diversity and Identity Separability:} 
Figure \ref{fig:sample_overview} presents samples of recent works that used synthetic datasets for \acrshort{fr} model training.
Synthetic face recognition models, SynFace \cite{Qiu2021} and USynthFace \cite{FBoutros2022USynthFace}, utilized synthetic images generated by DiscoFaceGAN \cite{Deng2020}. DiscoFaceGAN is based on disentangled representation learning to generate images from synthetic identities with predefined attributes e.g. pose, illumination, or expression. 
As generated images are explicitly controlled by a predefined set of attributes, such images might lack the intra-class diversity that exists in real-world face images, which is needed to successfully train \acrshort{fr} models. 
SFace \cite{Boutros2022SFace}, on the other hand, is a class-conditional GAN model that does not explicitly model these attributes. It is conditionally trained to generate synthetic images with a specific label. SFace provided images with a larger degree of intra-class variations, however, at the cost of less identity separability. 
In contrast to that, the DigiFace-1M \cite{DigiFace1M} images are produced by a \acrshort{3dmm} rendering process. The identities in DigiFace-1M are artificially defined as a combination of facial geometry, texture, and most notably hairstyle. Unfortunately, this approach is less suitable for research purposes, as it is extremely computationally expensive for generating a large dataset with an advanced computational rendering pipeline.

These observations are quantitatively supported by the genuine and imposter comparison score distribution plots in Figure \ref{fig:syn_vs_syn_distributios_overview} and corresponding verification performance results in Table \ref{tab:frm_training_validation_large_scale}.
The verification performance metrics include FMR100, and FMR1000, which are the lowest false non-match rate (FNMR) for a false match rate (FMR)$\leq1.0\%$ and $\leq0.1\%$, respectively, along with the Equal Error Rate (EER) \cite{iso_metric}. We additionally report the mean and standard deviation of the genuine and imposter scores. Further, we report the Fisher Discriminant Ratio (FDR) \cite{poh2004study} to provide an in-depth analysis of the separability of genuine and imposters scores. We used a pre-trained ElasticFace-Arc \cite{ElasticFace} by the corresponding author (model publicly available) to extract the feature embeddings of CASIA-WebFace \cite{Yi2014}, LFW \cite{LFWDatabase}, DiscoFaceGAN \cite{Deng2020} (used in SynFace \cite{Qiu2021} and USynthFace \cite{FBoutros2022USynthFace}), DigiFace-1M \cite{DigiFace1M}, and our \approachname for the investigations described in this subsection. CASIA-WebFace \cite{Yi2014} and LFW \cite{LFWDatabase} are authentic datasets, and they are commonly used to train or evaluate FR models \cite{ArcFace,ElasticFace}, respectively. We made the following observations:

\textit{1):} In comparison to the strong identity discrimination in the authentic LFW and CASIA-WebFace datasets, our \approachname (CPD of $0\%$) with two-stage and uniform identity-context sampling approach generates synthetic samples that clearly maintain identity discrimination, where, for example, the achieved EER on LFW was $0.002$ and by our models with CPD of $0\%$ were $0.007$ (Uniform) and $0.003$ (Two-Stage). It should be noted that CASIA-WebFace is an FR training dataset, and contains some noisy labels, as reported in \cite{DBLP:conf/iccv/WangWSWM19}, which contributes to the higher EER.  

\textit{2):} Training our proposed models with CPD demonstrates that with an increase of the \acrshort{cpd} probability, the intra-class diversity becomes larger as well, leading to more variation, in the head pose, the illumination, the expression, the accessories, and sometimes even in the age of the depicted individual, as shown in Figure \ref{fig:sample_overview}. This can be seen in the shift in the genuine distributions to left (i.e. genuine distribution shifts to the left as more challenging and realistic variations are generated) when the \acrshort{cpd} is increased, as shown in Figure \ref{fig:syn_vs_syn_distributios_overview} and the mean genuine score in Table \ref{tab:frm_training_validation_large_scale}. The higher the \acrshort{cpd} probability, the larger intra-class and realistic real-world variations can be achieved. It also comes at the cost of slightly losing identity discrimination. This observation can be seen in the EER value increase and the decrease in FDR values induced by incorporating CPD into the model training ( Table \ref{tab:related_models_metrics_table}). Hence, the probability of the \acrshort{cpd} can be interpreted as a trade-off between identity discrimination and intra-class diversity. Achieving a good balance for this trade-off is required to generate useful data for successful FR model training and to achieve high accuracies, as we will empirically test in the next section. 

\textit{3):} Among the recent SOTA synthetic FR training datasets, the synthetic dataset that is generated by DiscoFaceGAN \cite{Deng2020} (used in SynFace and USynthFace) maintains identity discrimination with EER of $0.01069$. However, it comes at the high cost of low intra-class diversity as the generated images are controlled by a predefined set of attributes, making it less optimal to train FR models. On the contrary, SFace possesses a large intra-class diversity, however, with a lower degree of identity discrimination in comparison to other synthetic datasets. DigiFace-1M maintains to some degree identity discrimination. However, it is obtained using a computationally expensive digital rendering process, leading to very high costs in the generation process. Our proposed \approachname with CPD achieved the best (most representative of reality and thus suitable for training) trade-off between identity discrimination and intra-class diversity and leads to SOTA synthetic-based FR, as we will present in the next section.

\textbf{Synthetic-based FR:}
We evaluate first the different variants of the proposed solution regarding their applicability to generate synthetic data for \acrshort{fr}. For that, we start by generating three sets of synthetic data using a uniform identity-context sampling, each with different CPD probabilities of $0$, $25\%$ and $50\%$. We also generated other three sets of synthetic data using a two-stage identity-context sampling, with the same three CPD probabilities. All training datasets contain $80{,}000$ samples ($5{,}000$ identities with $16$ samples each). We utilized ResNet-18 \cite{ResNet} with the CosFace \cite{CosFace} loss to train six FR models on each of our synthetic datasets.
For this small-scale FR model training, a fixed $40$ iteration over the entire training data, a step-based learning rate schedule with an initial rate of $0.1$ and reduces it by a factor of $0.1$ after the $22$nd, the $30$th, and the $35$th epoch, and an \acrshort{sgd} optimizer with $0.9$ momentum and $5e-4$ weight decay were used, following \cite{FBoutros2022USynthFace}. Evaluations are reported on the five benchmarks, LFW \cite{LFWDatabase}, AgeDb-30 \cite{AgeDB30Database}, CA-LFW \cite{CALFWDatabase}, CFP-FP \cite{CFPFPDatabase}, and CP-LFW \cite{CPLFWDatabase}, following their official evaluation protocols (see Table \ref{tab:frm_training_validation_small_scale}).

\tabcolsep=0.1cm
\begin{table}[t]
\centering
\ra{1.1}
\resizebox{\linewidth}{!}{%
    \begin{tabular}{@{}lccccccccccc@{}}\toprule
    \multicolumn{1}{c}{\textbf{ }}& \phantom{abc} & \multicolumn{3}{c}{\textbf{Operation Metrics}} & \phantom{abc} & \phantom{abc} & \multicolumn{5}{c}{\textbf{Score Distributions}}\\
    \cmidrule{7-12}
    \multicolumn{5}{c}{\textbf{ }}& \phantom{abc} & \multicolumn{2}{c}{genuine} &   \multicolumn{2}{c}{imposter} &  \multicolumn{1}{c}{\textbf{ }}\\
    
    \cmidrule{1-1} \cmidrule{3-5} \cmidrule{7-12} 
    \textbf{Method} && \acrshort{eer} $\downarrow$  & FMR100 $\downarrow$  & FMR1000 $\downarrow$  && mean & std & mean & std & \acrshort{fdr} $\uparrow$ \\
    \midrule
 CASIA-WebFace \cite{Yi2014} && 0.076  & 0.092  & 0.107120   &&  0.536 &  0.215  &  0.003  &  0.070  &  5.5409 \\

    LFW \cite{LFWDatabase} &&  0.002   &  0.002  &  0.002  &&  0.708  &  0.099  &  0.003  &  0.070  &  33.301 \\

    \cmidrule{1-1} \cmidrule{3-5} \cmidrule{7-12}
    
    DiscoFaceGAN \cite{Deng2020} \\ (SynFace \& USynthFace)  &&  0.011   &  0.011  &  0.051  &&  0.619  &  0.128  &  0.044  &  0.092  &  13.378 \\
    DigiFace-1M \cite{DigiFace1M, Wood2021}  &&  0.042   &  0.087  &  0.199  &&  0.512  &  0.140  &  0.099  &  0.084  &  6.372  \\
    SFace \cite{Boutros2022SFace}  &&  0.236  &  0.768  &  0.380  &&  0.159  &  0.125  &  0.016  &  0.079  &  0.941 \\
    
    \cmidrule{1-1} \cmidrule{3-5} \cmidrule{7-12}
    \textbf{\approachname} \\ \textbf{Synthetic Uniform (Ours)} &&&&&&&&&&& \\
    \midrule

    {\cacpdzeroshort} &&  \underline{0.007}   &  \underline{0.005}  &  \underline{0.019}  &&  0.528  &  0.117  &  0.014  &  0.069  &  \underline{14.243}  \\
    {\cacpdtwentyfiveshort} &&  0.130   &  0.385  &  0.607  &&  0.226  &  0.117 &  0.014  &  0.070  &  2.427 \\
    {\cacpdfiftyshort} &&  0.225  &  0.660  &  0.857  &&  0.149  &  0.108  &  0.013  &  0.070  &  1.120 \\
    
    \cmidrule{1-1} \cmidrule{3-5} \cmidrule{7-12}
    \textbf{w/ Synthetic Two-Stage (Ours)} &&&&&&&&&&& \\
    \midrule
    {\cacpdzeroshort} && \textbf{0.003}   &  \textbf{0.001}  &  \textbf{0.009}  &&  0.621  &  0.102  &  0.024  &  0.075  &  \textbf{22.172}  \\
    {\cacpdtwentyfiveshort}  &&  0.018  &  0.030  &  0.190  &&  0.448  &  0.114  &  0.023  &  0.075  &  9.733  \\
    {\cacpdfiftyshort}  &&  0.069  &  0.238  &  0.659  &&  0.309  &  0.122  &  0.021 &  0.075  &  4.064  \\
    
    \bottomrule
    \end{tabular}
}
\vspace{-3mm}
\caption{Evaluation of identity-separability in synthetic FR datasets proposed in the literature.
The first two rows present the results on authentic LFW and CASIA-WebFace datasets. These results are provided as a reference. In the next rows, we provide the evaluation results on SOTA synthetic FR datasets and our \approachnamewithoutspace.
Except for the first row, which shows the results obtained by computing the comparison scores on pre-defined pairs of authentic LFW, all the synthetic evaluations are based on a synthetically generated dataset with $5{,}000$ identities and $16$ sampled images per identity. The lowest errors and the highest genuine-imposter separability scores (FDR) on synthetic datasets are marked in \textbf{bold}. The second best per column is \underline{underlined}. }
\label{tab:related_models_metrics_table}
\vspace{-2mm}
\end{table}

\tabcolsep=0.04cm
\begin{table}\centering
\ra{1.0}
\resizebox{\linewidth}{!}{\begin{tabular}{@{}clccccccccccc@{}}\toprule
\multicolumn{2}{c}{\textbf{Training Dataset}}& \phantom{abc} & \multicolumn{9}{c}{\textbf{Verification Benchmarks $\uparrow$}}\\

\cmidrule{4-12}
\multicolumn{2}{c}{} & \phantom{abc} & \phantom{abc} & \phantom{abc} &\multicolumn{2}{c}{Cross-Age}  & \phantom{abc} & \multicolumn{2}{c}{Cross-Pose}  & \phantom{abc} & \phantom{abc}\\

\cmidrule{1-2} \cmidrule{4-4} \cmidrule{6-7}  \cmidrule{9-10} \cmidrule{12-12} 
Identity-context & model  && \acrshort{lfw} && \acrshort{agedb30}  & \acrshort{calfw} && \acrshort{cfpfp}  & \acrshort{cplfw}  && Average\\
\midrule
Uniform & \cacpdzeroshort &&  86.50 &&    60.30 &   72.13 && 62.26 &          62.40 && 68.72\\
{} & \cacpdtwentyfiveshort  &&  95.07 &&   74.87 &  \textbf{84.60} && 70.09 &    \textbf{73.08} && 79.54\\
{} & \cacpdfiftyshort &&  92.92 &&    71.43 &        80.48 && 66.79 &            69.03 && 76.13 \\

\cmidrule{1-2} \cmidrule{4-4} \cmidrule{6-7}  \cmidrule{9-10} \cmidrule{12-12} 
Two-Stage & \cacpdzeroshort &&  86.08 &&    61.50 &         71.68 &&  61.64 &          62.20 && 68.62\\
{} & \cacpdtwentyfiveshort &&  \underline{95.10} &&    \underline{76.18} &          \underline{84.17} &&  \underline{70.67} &         72.47 && \underline{79.72}\\
{} & \cacpdfiftyshort  &&  \textbf{95.42} &&    \textbf{76.55} &         83.22 && \textbf{70.90} &           \underline{72.57} && \textbf{79.73}\\

\bottomrule
\end{tabular}}
\vspace{-3mm}
\caption[Verification accuracies for small-scale \acrshort{frm} training on synthetic data]{Verification Accuracies (in $\%$) on five \acrshort{fr} benchmarks for \acrshort{frm}s (ResNet-18) trained  on $80{,}000$ samples ($5{,}000$ identities with $16$ images per identity) generated with different levels of \acrshort{cpd} probability. The best  accuracies are marked in \textbf{bold} and the second best is \underline{underlined}.}
\label{tab:frm_training_validation_small_scale}
\vspace{-6mm}
\end{table}

It can be observed from Table \ref{tab:frm_training_validation_small_scale} that FR models trained on our \approachname datasets achieved high verification accuracies, even only using a small synthetic dataset ($80$K samples) for training.
For the models trained with datasets generated by \approachname with uniform identity-context sampling, the best verification accuracies were achieved using the FR model trained on the dataset generated by \approachname with CPD25 ($79.54\%$ average accuracies). For the models trained with datasets generated by \approachname with two-stage identity-context sampling, both FR models trained on  the datasets generated by \approachname with CPD25 and CPD50 achieved very competitive results. The best overall verification accuracy was achieved by \approachname using two-stage identity-context sampling with CPD50 (average accuracies of $79.73\%$) which is very close to two-stage identity-context sampling with CPD25 (average accuracies of $79.72\%$) and uniform identity-context sampling and CPD25 (Average accuracies of $79.54\%$). 
Although the CPD approach slightly affects, to some degree, the identity discrimination in the synthetically generated samples (as we discussed in the previous section), it leads to significant improvements in the FR verification accuracies as it increases the intra-class variations in the generated samples. As we mentioned in the previous section, achieving a realistic trade-off between identity discrimination and intra-class diversity is required for the FR training dataset to achieve high verification accuracies on evaluation benchmarks.

\tabcolsep=0.04cm
\begin{table*}[!]\centering
\ra{1.0}
\resizebox{0.95\linewidth}{!}{%
\begin{tabular}{@{}lccccrccccccccccccc@{}}\toprule
\multicolumn{2}{c}{\textbf{Method}} & \multicolumn{3}{c}{\textbf{Dataset}} \phantom{abc} &  \phantom{abc} & \multicolumn{9}{c}{\textbf{Verification Benchmarks $\uparrow$}}\\

\cmidrule{8-15}

\multicolumn{2}{c}{}&  \multicolumn{3}{c}{} & \phantom{abc} & \phantom{abc} &  \phantom{abc} &\multicolumn{2}{c}{Cross-Age}  & \phantom{abc} & \multicolumn{2}{c}{Cross-Pose} & \phantom{abc} & \phantom{abc}\\
\hline

name & Aug. & Id & \shortstack{$N$ per Id} & images  && \acrshort{lfw} && \acrshort{agedb30}  & \acrshort{calfw}  && \acrshort{cfpfp}  & \acrshort{cplfw}  && Avg.\\
\midrule

\rowcolor{Gray}
ElasticFace \cite{ElasticFace} (CASIA-WebFace \cite{Yi2014}) & \FeatureFalse & 10.5 K  &  -  &  494 K &&  99.52  &&  94.77  &   93.93  &&    95.52 & 90.38 && 94.82 \\
\rowcolor{Gray}
ElasticFace \cite{ElasticFace} (MS1MV2 \cite{MSCeleb1MDataset}) & \FeatureFalse & 85 K  &  -  &  5{,}800 K &&  99.82  &&  98.27  &   96.03  &&    98.61 & 93.17 && 97.18 \\

\midrule
SynFace \cite{Qiu2021} & \FeatureFalse & 10 K  &  50  &  500 K &&  91.93  &&  61.63  &  74.73  &&  75.03  &  70.43 && 74.75\\

\hline
SFace \cite{Boutros2022SFace} & \FeatureFalse & 10.5 K  &  10  &  105 K &&  87.13  &&  63.30  &  73.47  &&  68.84  &  66.82 && 71.91\\

{} & \FeatureFalse & 10.5 K &  20  &  211 K &&  90.50  && 69.17  &  76.35  &&  73.33  &  71.17 && 76.10\\

{} & \FeatureFalse & 10.5 K  &  40  &  423 K &&  91.43  &&  69.87  &  76.92  &&  73.10  &  73.42 && 76.95\\

{} & \FeatureFalse &  10.5 K  &  60  &  634 K &&  91.87  &&  71.68  &  77.93  &&  73.86  &  73.20 && 77.71 \\ \hline

USynthFace \cite{FBoutros2022USynthFace} + RandAugment \cite{FBoutros2022USynthFace}  & \FeatureTrue &  100 K  &  1  &  100 K &&  92.12  &&  71.08  &  76.15  &&  78.19  &  71.95 && 77.90\\
USynthFace \cite{FBoutros2022USynthFace}     + RandAugment \cite{FBoutros2022USynthFace}  & \FeatureTrue & 200 K  &  1  &  200 K &&  91.93  &&  71.23  &  76.73  &&  78.03  &  72.27 && 78.04\\
USynthFace \cite{FBoutros2022USynthFace} + RandAugment \cite{FBoutros2022USynthFace}  & \FeatureTrue & 400 K &  1  &  400 K &&  92.23  &&  71.62  &  77.05  &&  78.56  &  72.03 && 79.30\\ \hline
DigiFace-1M \cite{DigiFace1M} & \FeatureFalse  &10 K  &  50  &  500 K && 88.07  && 60.92  & 69.23 && 70.99   & 66.73 && 71.19 \\
\multicolumn{1}{r}{+ Augmentation\cite{DigiFace1M} }  & \FeatureTrue  &10 K  &  50  &  500 K &&  95.40  &&  76.97  &  78.62  &&  \textbf{87.40}  &  78.87 && 83.45\\
\hline
ExFaceGAN(SG3) \cite{ExFaceGAN} + RandAugment \cite{FBoutros2022USynthFace} & \FeatureTrue  &10 K  &  50  &  500 K &&  90.47              && 72.85      & 78.60         && 72.70                        & 69.27          && 76.78             \\
ExFaceGAN(Con) \cite{ExFaceGAN} + RandAugment \cite{FBoutros2022USynthFace}  & \FeatureTrue  &10 K  &  50  &  500 K && 93.50 && 78.92  & 82.98 && 73.84                 & 71.60          && 80.17  \\
\hline
IDnet \cite{IDnet} & \FeatureFalse  &10.5 K  &  50  &  528 K &&  84.83              && 63.58          & 71.50         && 70.43                              & 67.35                && 71.54   \\
\multicolumn{1}{r}{+ RandAugment \cite{FBoutros2022USynthFace} }   & \FeatureTrue  &10.5 K  &  50  &  528 K && 92.58             && 73.53     & 79.90               && 75.40                         & 74.25                && 79.13  \\
\midrule

\approachname
\cacpdtwentyfiveshort (Uniform) - Ours & \FeatureFalse & 5 K  &  16  &  80 K &&  94.37  &&  76.70  &   85.02  &&  70.31  &  72.18 && 79.72  \\
{} & \FeatureFalse & 5 K &  32  &  160 K &&  96.23  &&   80.50  &   88.30  &&    73.87  &    75.42   && 82.86\\
{} & \FeatureFalse & 10 K &  16  &  160 K && 96.72 &&     81.52  & 88.83  &&  75.43   &      76.85   && 83.87\\
{} & \FeatureFalse & 10 K  &  50  &  500 K &&  97.68  &&  \underline{84.63}  &  \underline{90.58}  && 82.39  &  79.70  && \underline{87.00}\\
\multicolumn{1}{r}{+ RandAugment \cite{FBoutros2022USynthFace} } & \FeatureTrue & 10 K  &  50  &  500 K &&  \textbf{98.00}  &&   \textbf{86.43}  &   \textbf{90.65}  &&  \underline{85.47}  &   \textbf{80.45} && \textbf{88.20} \\ \hline

\approachname
\cacpdtwentyfiveshort (Two-Stage) - Ours & \FeatureFalse & 5 K &  16  &  80 K &&  95.27  &&  75.72  &  84.63  &&  70.07  & 72.65 && 79.67\\
{} & \FeatureFalse & 5 K  &  32  &  160 K &&  95.92  &&  77.85  &  85.40  &&  72.97  & 74.18 && 81.26\\
{} & \FeatureFalse  & 10 K &  16  &  160 K &&  96.60  &&  79.87  &  87.03  &&  74.47  & 75.47 && 82.69\\
{} & \FeatureFalse  & 10 K &  50  &  500 K &&  97.52  &&  81.65  &  87.77  &&  80.03  & 77.60 && 84.91 \\

\multicolumn{1}{r}{+ RandAugment \cite{FBoutros2022USynthFace} } & \FeatureTrue  & 10 K &  50  &  500 K &&  97.32  &&  83.45  &  87.98  &&  81.80  & 77.92 && 85.69 \\
\hline
\approachname \cacpdfiftyshort (Two-Stage) - Ours & \FeatureFalse & 5 K  &  16  &  80 K &&  95.58  &&     74.00  &   83.57  &&          70.43  &           72.53 && 79.22 \\
{} & \FeatureFalse & 5 K  &  32  &  160 K &&  96.40  &&     78.23  &   85.87  &&          73.00  &    75.38      && 81.78   \\
{} & \FeatureFalse  & 10 K &  16  &  160 K &&  97.00  &&     80.85  &   86.38  &&          74.24  &   76.73  && 83.04        \\
{} & \FeatureFalse  & 10 K &  50  &  500 K &&  97.87  &&  83.53  &   89.05  &&          80.00  &  78.95  && 85.88\\

\multicolumn{1}{r}{+ RandAugment \cite{FBoutros2022USynthFace} } & \FeatureTrue & 10 K &  50  &  500 K && \underline{97.97}  &&  84.40  &  88.52  &&  83.87  & \underline{79.88}  && 86.93\\

\bottomrule

\end{tabular}
}
\vspace{-2mm}
\caption{Verification Accuracies (in $\%$) on five \acrshort{fr} benchmarks for  \acrshort{sota} synthetic-based \acrshort{fr}s. the first two rows present the results of \acrshort{fr}s trained on authentic data. These results are provided as references. 
All results of previous work are copied from their corresponding works.
The synthetic-based \acrshort{fr} utilized ResNet-50 as network architecture. 
The best verification accuracies of synthetic-based \acrshort{fr} are marked in \textbf{bold} and the second best are \underline{underlined}. }
\label{tab:frm_training_validation_large_scale}
\vspace{-6mm}
\end{table*}

\textbf{Comparison to the SOTA Synthetic-based FR:}
We compare the verification performance of FR models trained with our synthetic data with the recent works that proposed the use of synthetic data for FR training. To provide a fair comparison, we followed the SOTA synthetic FR approaches \cite{Qiu2021,Boutros2022SFace,FBoutros2022USynthFace} to generate $500{,}000$ images from $10{,}000$ synthetic identities with $50$ samples each, and utilized these images to train ResNet-50 \cite{ResNet}. We also followed the exact training setups described in \cite{FBoutros2022USynthFace} to train $15$ FR models using different dataset sizes, three levels of CPD, and two identity-context sampling approaches to investigate the effect of training dataset width and depth and identity-context sampling mechanisms on \acrshort{fr} verification accuracy. The achieved results by SOTA synthetic-based FR and our proposed \approachname are presented in Table \ref{tab:frm_training_validation_large_scale}. One can observe the following from the reported results in Table \ref{tab:frm_training_validation_large_scale}:

\textit{1):} With a clear margin, FR models trained with our \approachname outperformed all previous synthetic-based FRs. The best achieved average accuracy by our models was $88.20\%$ and the average accuracy of SOTA synthetic-based FR was $83.45\%$ (achieved by DigiFace-1M \cite{DigiFace1M}).

\textit{2):} Increasing our training dataset size did increase the FR verification accuracies in all experimental settings. This indicates that increasing our synthetic training dataset size could further improve the FR verification accuracy.

\textit{3):} Increasing our dataset width (the number of identities) led to higher accuracies in comparison to the case when the dataset depth (the number of images per identity) is increased. For example, \approachname with uniform sampling (CPD25) led to an average accuracy of $82.86\%$ when using 160K samples (5K identities with 32 images per identity). This accuracy improved to $83.87\%$ when we trained on 160K (10K images with 16 images per identity). This observation can be concluded from all of our experiments. 

\textit{4):} Introducing data augmentation \cite{FBoutros2022USynthFace} to model training improved the verification accuracies, where we additionally trained the best model in each setting with RandAugment \cite{RandAugment} using settings provided in \cite{FBoutros2022USynthFace}. This experiment is highlighted with \textit{+ RandAugment(4, 16)}. It should be noted that previous SOTA synthetic-based FR models (DigiFace-1M \cite{DigiFace1M}, ExFaceGAN \cite{ExFaceGAN}, IDnet \cite{IDnet} and USynthFace \cite{FBoutros2022USynthFace}) also applied extensive data augmentation in their model training.

\textit{5):} With only $160$K training samples ($10$K identities with $16$ images each), our approach outperformed all previous synthetic-based FR that used GAN-generated training face images (SynFace, SFace, ExFaceGAN, IDnet and USynthFace) and the one that used digital face rendering (DigiFace-1M), where the average accuracy by our best model trained on $160$K images was $83.87\%$ and the average accuracies by SynFace ($500$K training samples), SFace ($634$K training samples), USynthFace ($400$K training samples) and DigiFace-1M ($500$K training samples) were $74.75\%$, $77.71\%$, $79.30\%$ and $83.45\%$, respectively.

\textit{6):} Although it is out of our paper comparison scope, we compared our synthetic-based FRs with the SOTA FRs trained on authentic datasets (CASIA-WebFace and MS1MV2).  Using only our synthetic data, our proposed models achieved very competitive results to models trained on the large authentic training dataset. This clearly indicates that synthetically generated data is growing to become a valid alternative to authentic data to train FR. We achieved for example $98.00\%$ accuracy on LFW \cite{LFWDatabase}, which is competitive to SOTA FR accuracies that are trained on authentic datasets, $99.52\%$ using CASIA-WebFace and $99.82\%$ using MS1MV2. This achieved accuracy ($98.00\%$) on LFW is even higher than human-level performance in face verification on LFW ($97.5\%$ \cite{DBLP:conf/iccv/KumarBBN09}).

\vspace{-3mm}
\section{Conclusion}
\vspace{-1mm}
An ideal training dataset for \acrshort{fr} would strongly ensure identity discrimination and exhibit a realistic and large intra-class variation. Authentic face datasets proposed in the literature hold, to a large degree, these properties, leading to breakthroughs in verification accuracy. However, future developments using these authentic datasets might be infeasible due to increased legal and ethical concerns about the use and distribution of sensitive authentic data for \acrshort{fr} development. In this work, we proposed \approachnamewithoutspace, an identity-conditioned generative model based on a \acrshort{dm} to generate synthetic identity-specific face images. We also proposed \acrshort{cpd} as a simple, yet effective mechanism to prevent the model from overfitting to the identity context and to control the trade-off between identity-separability and intra-class variation.
Utilizing our synthetically generated data for \acrshort{fr} training led to new \acrshort{sota} verification accuracies on five mainstream \acrshort{fr} benchmarks, outperforming all recent \acrshort{sota} synthetic-based \acrshort{fr} approaches.

\textbf{Acknowledgment}
This research work has been funded by the German Federal Ministry of Education and Research and the Hessian Ministry of Higher Education, Research, Science and the Arts within their joint support of the National Research Center for Applied Cybersecurity ATHENE. This work has been partially funded by the German Federal Ministry of Education and Research (BMBF) through the Software Campus Project.

{\small
\bibliographystyle{ieee_fullname}
\bibliography{main}
}

\end{document}